\documentclass{article}

    \PassOptionsToPackage{numbers, compress}{natbib}

\usepackage[preprint]{neurips_2023}

\usepackage[utf8]{inputenc} 
\usepackage[T1]{fontenc}    
\usepackage{hyperref}       
\usepackage{url}            
\usepackage{booktabs}       
\usepackage{amsfonts}       
\usepackage{nicefrac}       
\usepackage{microtype}      
\usepackage{xcolor,wrapfig}         
\usepackage{graphicx}
\usepackage{makecell}
\usepackage{bbm}

\usepackage{amsmath}
\usepackage{amsfonts}
\usepackage{amssymb}
\usepackage{xspace}

\usepackage{caption}
\usepackage{subcaption}

\usepackage{hyperref}

\newcommand{\ie}{\textit{i.e.}}

\newcommand{\xbf}{\mathbf{x}\xspace}

\newcommand{\ebf}{\mathbf{e}\xspace}

\newcommand{\Xbf}{\mathbf{X}\xspace}
\newcommand{\Zbf}{\mathbf{Z}\xspace}

\newcommand{\Abf}{\mathbf{A}\xspace}

\newcommand{\CG}{G\xspace}
\newcommand{\CV}{V\xspace}
\newcommand{\CT}{\mathcal{T}\xspace}
\newcommand{\CE}{E\xspace}

\newcommand{\CD}{\mathcal{D}\xspace}
\newcommand{\CX}{\mathcal{X}\xspace}
\newcommand{\CY}{\mathcal{Y}\xspace}

\newcommand{\CM}{\mathcal{M}\xspace}

\newtheorem{prob}{Problem}

\title{A Survey on Explainability of Graph Neural Networks}
\author{
Jaykumar Kakkad\thanks{Both authors contributed equally.},   Jaspal Jannu$^*$ \\
  University of Illinois Chicago, USA \\
  \texttt{\{jkakka4,jjannu2\}@uic.edu} \\
  \And
  Kartik Sharma \\
  Georgia Institute of Technology, USA \\
  \texttt{ksartik@gatech.edu} \\
  \And
  Charu Aggarwal \\
  IBM T. J. Watson Research Center, USA \\
  \texttt{charu@us.ibm.com} \\
  \And
  Sourav Medya \\
  University of Illinois, Chicago, USA\\
  \texttt{medya@uic.edu} \\
}

\begin{document}

\maketitle 
 
\begin{abstract}

Graph neural networks (GNNs) are powerful graph-based deep-learning models that have gained significant attention and demonstrated remarkable performance in various domains, including natural language processing, drug discovery, and recommendation systems. However, combining feature information and combinatorial graph structures has led to complex non-linear GNN models. Consequently, this has increased the challenges of understanding the workings of GNNs and the underlying reasons behind their predictions. 
To address this, numerous explainability methods have been proposed to shed light on the inner mechanism of the GNNs. Explainable GNNs improve their security and enhance trust in their recommendations. This survey aims to provide a comprehensive overview of the existing explainability techniques for GNNs. 
We create a novel taxonomy and hierarchy to categorize these methods based on their objective and methodology. 
We also discuss the strengths, limitations, and application scenarios of each category. Furthermore, we highlight the key evaluation metrics and datasets commonly used to assess the explainability of GNNs. This survey aims to assist researchers and practitioners in understanding the existing landscape of explainability methods, identifying gaps, and fostering further advancements in interpretable graph-based machine learning.

\end{abstract}
\section{Introduction}
Recent years have seen a tremendous rise in the use of Graph Neural Networks (GNNs) for real-world applications, ranging from healthcare~\cite{zitnik2018modeling,zitnik2019machine}, drug design~\cite{xiong2021graph,drug-design, drug-ligand}, recommender systems~\cite{chen2022grease}, and fraud detection~\cite{fraud-detect}. Predictions made in these domains have a substantial impact and therefore require to be highly trustworthy. In the realm of deep learning, one effective approach to enhance trust in these predictions is to provide an explanation supporting them~\cite{trust-AI}. These explanations elucidate the model's predictions for human understanding and can be generated through various methods. For instance, they may involve identifying important substructures within the input data~\cite{pgexplainer, Graph-mask, subgraphX}, providing additional examples from the training data~\cite{SE-GNN}, or constructing counterfactual examples by perturbing the input to produce a different prediction outcome~\cite{cfgnnex, cf^2-counter, chen2022grease}.

The interpretability of deep learning models is influenced by the characteristics of the input domain as the content and the complexity of explanations can vary depending on the inputs. When it comes to explaining predictions made by graph neural networks (GNNs), several challenges arise. First, since graphs are combinatorial data structures, finding important substructures by evaluating different combinations that maximize a certain prediction becomes difficult. Second, attributed graphs contain both node attributes and edge connectivity, which can influence the predictions and they should be considered together in explanations. Third, explanations must be adaptable to different existing GNN architectures. Lastly, explanations for the local tasks (e.g., node or edge level) may differ from those for global tasks (e.g., graph level). Due to these challenges, explaining graph neural networks is non-trivial and a large variety of methods have been proposed in the literature to tackle it~\cite{ying2019gnnexplainer, GSAT, D_invariant_rationale, Excitation-BP, guided-bp, pgm-ex, pgexplainer, xgnn, cf^2-counter, robust-counter, meg-counter}. With the increasing use of GNNs in critical applications such as healthcare and recommender systems, and the consequent rise in their explainability methods, we provide an updated survey of the explainability of GNNs. Additionally, we propose a novel taxonomy that categorizes the explainability methods for GNNs, providing a comprehensive overview of the field.

Existing surveys on GNN explainability predominantly focus on either factual methods \cite{First-survey,survey2-prfe,survey5-li} or counterfactual methods \cite{survey-counter} but not both. 
These surveys thus lack a comprehensive overview of the different methods in the literature by limiting the discussion to specific methods~\cite{First-survey,survey5-li} or only discussing them under the broad umbrella of trustworthy GNNs~\cite{survey2-prfe, survey3-gl}.
Our survey aims to bridge this gap by providing a comprehensive and detailed summary of existing explainability methods for GNNs. We include both factual as well as counterfactual methods of explainability in GNNs. To enhance clarity and organization, we introduce a novel taxonomy to categorize these methods for a more systematic understanding of their nuances and characteristics. 

\subsection{Graph Neural Networks}
Graph neural networks (GNNs) have been used to learn powerful representations of graphs~\cite{kipf2016semi,velivckovic2017graph,hamilton2017inductive}. Consider a graph $\CG$ given by $\CG = (\CV, \CE)$, where $\CV$ denotes the set of $n$ nodes and $\CE$ denotes the set of $m$ edges. We can create an adjacency matrix $\Abf \in {[0, 1]}^{n \times n}$ such that $A_{ij} = 1$ if $(i,j) \in \CE$ and $0$ otherwise. Each node may have attributes, given by the matrix $\Xbf \in \mathbb{R}^{n \times F}$ such that each row $i$ stores the $F$-dimensional attribute vector for node $i$. A GNN model $\CM$ embeds each node $v \in \CV$ into a low-dimensional space $\Zbf: \mathbb{R}^{n \times d}$ by following this message passing rule for $k$ steps as

\begin{equation}
    \Zbf_v^{(k+1)} = \textsc{Update}_{\Phi}(\Zbf_v^{(k)}, \textsc{Agg}(\{\textsc{Msg}_{\Theta}(\Zbf_v^{(k)}, \Zbf_u^{(k)}): (u, v) \in \CE\})),
\end{equation}

such that $\Zbf^{0} = \Xbf$ and $\Zbf := \Zbf^{(k)}$. Different instances of the update $\textsc{Update}_{\Phi}$, aggregation $\textsc{Agg}_{\Phi}$ and message generator $\textsc{Msg}_{\Theta}$ functions give rise to different GNN architectures. For example, GCN~\cite{kipf2016semi} has an identity message, a mean aggregation, and weighted update functions, while GAT~\cite{velivckovic2017graph} learns an attention-based message generation instead. These embeddings are trained for a specific task $\CT$ that can be either supervised (e.g., node classification, graph classification, etc.) or unsupervised (e.g., self-supervised link prediction, clustering, etc.). 

\subsection{Explainability in ML}
\begin{prob}[Explainability~\cite{burkart2021survey}]
    Consider a supervised task $\CT$ with the aim of learning a mapping from $\CX$ to $\CY$, and a model $\CM$ trained for this task. Given a set of $(\xbf, y)$ pairs $\subseteq (\CX, \CY)$ and the model $\CM$, generate an explanation $\ebf$ from a given set $\CD_{E}$ such that $\ebf$ ``explains'' the prediction $\hat{y} = \CM(\xbf)$. 
\end{prob}

These explanations can be either \textit{local} to a single test input $(\xbf, y)$ or \textit{global} when they explain prediction over a specific dataset $\CD' \subseteq (\CX, \CY)$. Further, the explanation can be generated either \textit{post-hoc} (\ie, after the model training) or \textit{ante-hoc} where the model itself is \textit{self-interpretable}, \ie, it explains its predictions. With some exceptions, post-hoc explanations usually consider a black-box access to the model while self-interpretable methods update the model architecture and/or training itself. We can further differentiate the explanation methods based on their content, \ie, the explanation set $\CD_E$. \textit{Local explanations} only consider the local neighborhood of the given data instance while \textit{global explanations} are concerned about the model's overall behavior and thus, searches for patterns in the model's predictions. On the other hand, explanations can also be \textit{counterfactual}, where the aim is to explain a prediction by providing a contrasting example that changes it.

\section{Overview}


With the widespread adoption of GNNs across various applications, the demand for explaining their predictions has grown substantially. Moreover the GNN-based models are becoming more complex \cite{meta_mandal2023}. Recently, the community has witnessed a surge in efforts dedicated to the explainability of GNNs. These methods exhibit variations in terms of explanation types, utilization of model information, and training procedures, among other factors. We organize and categorize these methods to develop a deeper understanding of the existing works and provide a broad picture of their applicability in different scenarios.

\begin{figure*}[htbp]
  \centering
  \vspace{-.5mm}
  \includegraphics[width= .95\textwidth]{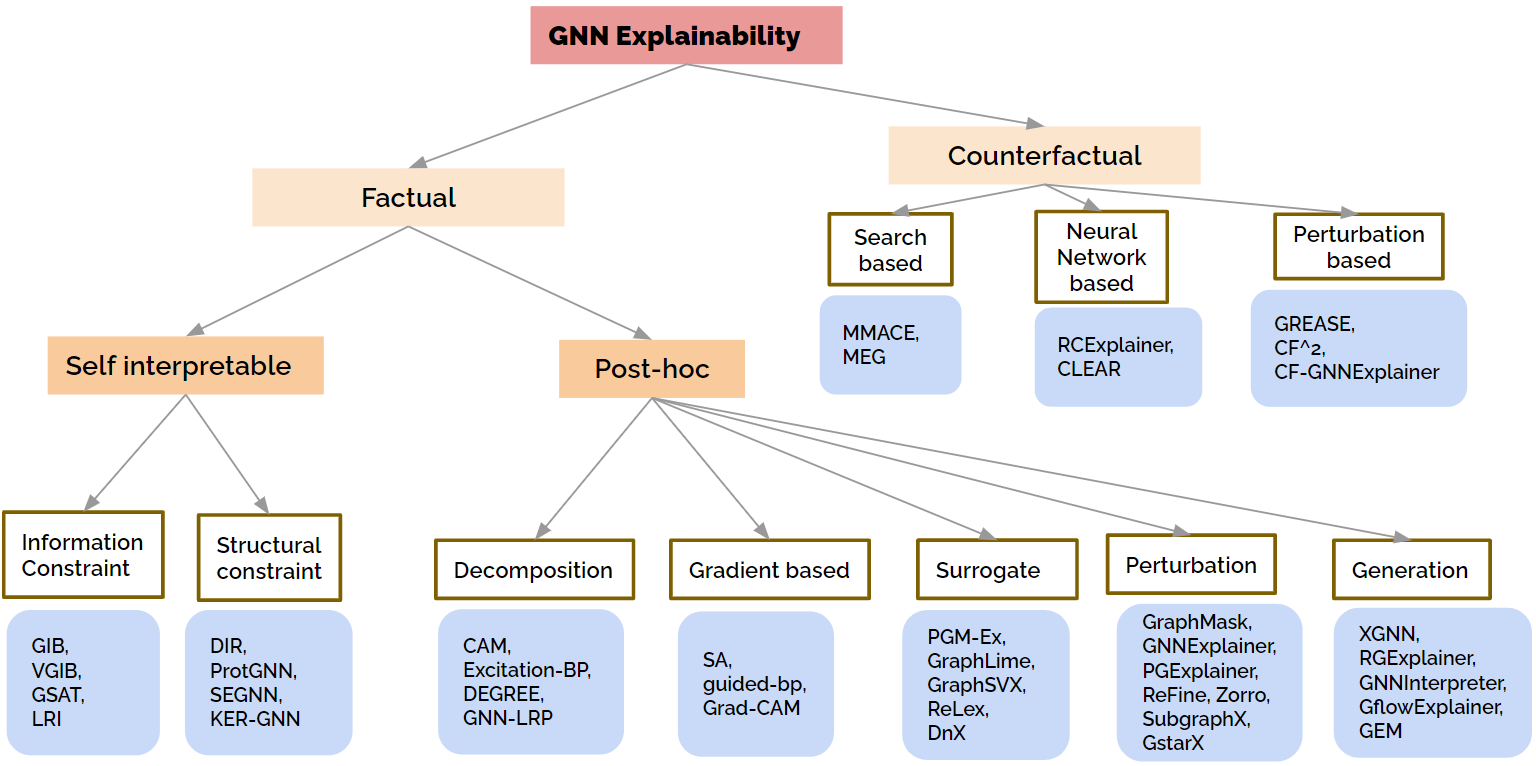}
  \centering
  \caption{\small \textbf{Overview of the Schema.} \textbf{(1) Factual.}  Information constraints: GIB \cite{GIB}, VGIB \cite{VGIB}, GSAT \cite{GSAT}, LRI \cite{inject-explain}; Structural Constraints: DIR \cite{D_invariant_rationale}, ProtGNN \cite{protgnn}, SEGNN \cite{SE-GNN}, KER-GNN \cite{kergnns}; Decomposition: CAM \cite{Excitation-BP}, Excitation-BP \cite{Excitation-BP}, DEGREE \cite{degree}, GNN-LRP \cite{GNN-LRP}; Gradient-based: SA \cite{guided-bp} , Guided-BP \cite{guided-bp} , Grad-CAM \cite{Excitation-BP}; Surrogate: PGM-Ex \cite{pgexplainer}, GraphLime \cite{graphlime}, GraphSVX \cite{graphsvx}, ReLex \cite{RELex}, DnX \cite{distilexplain}; Perturbation-based: GNNExplainer\cite{ying2019gnnexplainer}, GraphMask \cite{Graph-mask}, PGExplainer \cite{pgexplainer}, ReFine \cite{ReFine}, ZORRO \cite{zorro}, SubgraphX \cite{subgraphX}, GstarX \cite{gstarx}; Generation: XGNN \cite{xgnn}, RGExplainer \cite{RL-enhanced}, GNNInterpreter \cite{gnninterpreter}, GFlowExplainer \cite{Gflow}, GEM \cite{Gen-causal}; \textbf{(2) Counterfactual.} Search-based: MMACE \cite{agnostic-counter} , MEG \cite{meg-counter}; Neural Network-based: RCExplainer \cite{robust-counter}, CLEAR \cite{clear-counter}; Perturbation-based: GREASE \cite{chen2022grease}, CF2 \cite{cf^2-counter}, CF-GNNexplainer \cite{cfgnnex}
  }
  \label{fig:schema}
\end{figure*}

\noindent
\textbf{Main Schema: Factual and Counterfactual Methods. } Figure \ref{fig:schema} provides an overview of the broad categorization of the existing works. Based on the type of explanations, we first make two broad categories: (1) Factual and (2) Counterfactual. Factual methods aim to
find an explanation in the form of input features with the maximum influence over the prediction. These explanations can be a set of either node features or a substructure (set of nodes/edges) or both. On the other hand, counterfactual methods provide an explanation by finding the smallest change in the input graph that changes the model's prediction. Hence, counterfactual explanations can be used to find a set of similar features that can alter the prediction of the model.\\
\noindent
\textbf{Organization. }In the following sections, we describe each category in detail and provide summary of various explainability methods in each category. In Sec. \ref{sec::Factual}, we describe the factual approaches which are further classified into self-interpretable and post-hoc categories. In Sec. \ref{sec::Counterfactual}, the counterfactual methods are categorized into perturbation-based, neural network-based and search-based methods. Sec. \ref{sec::Others} presents three special categories of explainers such as temporal, global and causality-based. In Sec. \ref{sec::application}, we overview the explainer methods that are relevant for specific applications in different domains such as in social networks, biology, and computer security. Lastly, we review widely used datasets in Sec. \ref{sec::datasets} and evaluation metrics in Sec. \ref{sec::eval}.

\section{Factual}
\label{sec::Factual}
We classify the factual explainer methods broadly into two categories based on the nature of the integration of the explainability architecture with the main model as follows.


\begin{figure}[t]
\vspace{-4mm}
     \centering
     \begin{subfigure}[b]{0.48\textwidth}
         \centering
         \includegraphics[width=\textwidth]{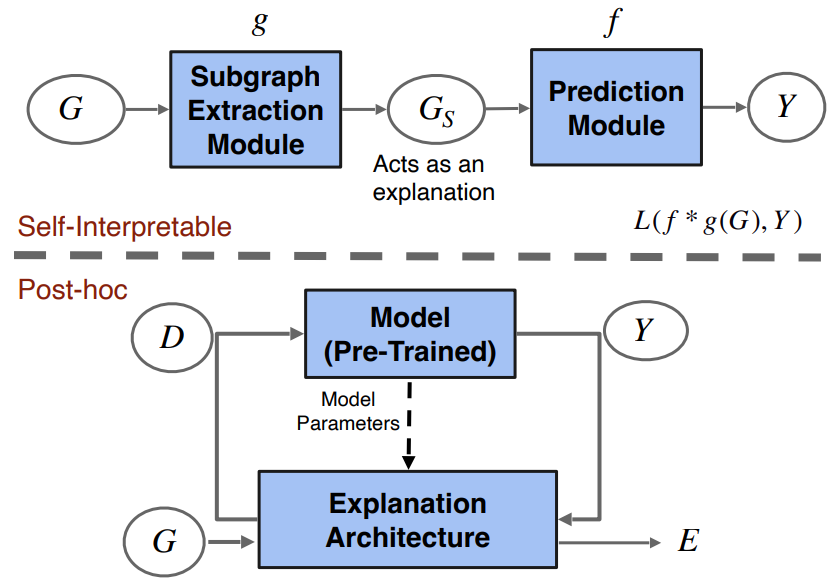}
         \caption{Self-Interpretable \& Post-hoc}
         \label{fig:SE_vs_PH}
     \end{subfigure}
     \hfill
     \begin{subfigure}[b]{0.5\textwidth}
         \centering
         \includegraphics[width=\textwidth]{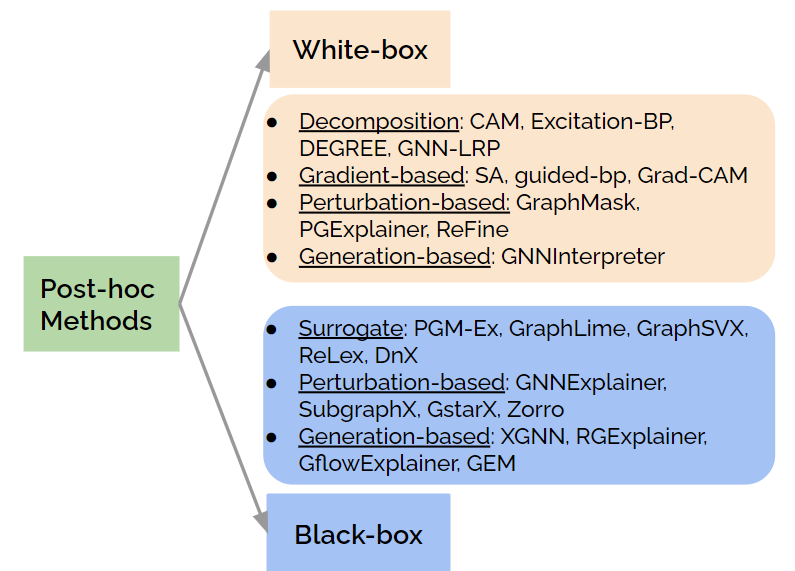}
         \caption{Categorization of Post-hoc methods}
         \label{fig:white_black_box}
     \end{subfigure}
        \caption{(a) \small \textbf{Self-interpretable and post-hoc architectures :} In self-interpretable methods, the subgraph extraction module \(g\) uses constraints to find an informative subgraph \(\CG_s\) from the input graph \(\CG\). The prediction module \(f\) uses this \(\CG_s\) to predict the label \(Y\). In contrast, Post-hoc methods consider model as pre-trained with fixed weights. For any instance \(\CG\), post-hoc methods generate explanation using model's input \(\CD\), output \(Y\) and in some cases the model's internal parameters. (b) \small \textbf{White-box and Black-box 
        post-hoc methods:} Methods are shown in the individual categories. Decomposition-based: CAM \cite{Excitation-BP}, Excitation-BP \cite{Excitation-BP}, DEGREE \cite{degree}, GNN-LRP \cite{GNN-LRP}; Gradient-based: SA \cite{guided-bp} , Guided-BP \cite{guided-bp} , Grad-CAM \cite{Excitation-BP}; Surrogate: PGM-Ex \cite{pgexplainer}, GraphLime \cite{graphlime}, GraphSVX \cite{graphsvx}, ReLex \cite{RELex}, DnX \cite{distilexplain}; Perturbation-based: GNNExplainer~\cite{ying2019gnnexplainer}, GraphMask \cite{Graph-mask}, PGExplainer \cite{pgexplainer}, ReFine \cite{ReFine}, ZORRO \cite{zorro}, SubgraphX \cite{subgraphX}, GstarX \cite{gstarx}; Generation-based: XGNN \cite{xgnn}, RGExplainer \cite{RL-enhanced}, GNNInterpreter \cite{gnninterpreter}, GFlowExplainer \cite{Gflow}, GEM \cite{Gen-causal}.}
        \label{}
\end{figure}


\begin{itemize}
\item \textbf{Post-hoc:} Post-hoc methods do not have the explainable architecture inbuilt into the model to attribute a model's prediction to the input. As seen in Figure \ref{fig:SE_vs_PH}, the explainability architecture (EA) is separated from the model which is pre-trained with fixed weights. For any instance \(\CG\), post-hoc methods generate an explanation using the model's input \(\CD\), output \(Y\) and sometimes even internal parameters of the model. Note that different EAs use different inputs \(\CD\) that are fed to the model. Post-hoc methods might not be always accurate as they may end up extracting features that are spuriously correlated with the task \cite{protgnn, not-posthoc, GSAT,kosan2023robust}.

\item \textbf{Self-interpretable:} Contrary to post-hoc methods, self-interpretable explainable methods design explainability architecture directly inside the model. As seen in Figure \ref{fig:SE_vs_PH}, these methods usually have two modules. The subgraph extraction module (the function \(g\)) uses constraints to find an informative subgraph \(\CG_s\) from the input graph \(\CG\). Then, the prediction module \(f\) uses \(\CG_s\) to predict label \(Y\). \(\CG_s\) also acts as an explanation. Both modules are trained together with an objective  \(L(f \circ g(\CG), Y)\) to minimize the loss between prediction \(f \circ g (\CG)\) and label $Y$. One major drawback of self-interpretable models is that the good interpretability of the models is often at the cost of the prediction accuracy \cite{GSAT}.


\end{itemize}

\subsection{Post-hoc}
\label{sec:post_hoc}
We divide the post-hoc methods based on their approaches used to find explanation into the following categories: a) Decomposition-based methods (Sec. \ref{sec:white_decom}), b) Gradient-based methods (Sec. \ref{sec::gradient-based}), c) Surrogate methods (\ref{sec::surrogate_methods}), d) Perturbation-based methods (Sec. \ref{sec::perturbation}), e) Generation-based methods (\ref{sec::generation-based}). The post-hoc methods can also be categorized based on their requirement to access the internal parameters of the model. As seen in figure \ref{fig:white_black_box}, this division results into the following categories of the methods: \textbf{white-box} and \textbf{black-box}. \\
\noindent
\textbf{White-box}: These methods require access to internal model parameters or embeddings to provide explanations. For instance, all decomposition-based methods (Sec. \ref{sec:white_decom}) require model parameters such as node weights of each layer to compute an importance score of different parts of the input. Even gradient based methods (Sec. \ref{sec::gradient-based}) require access to the gradients. Thus, all methods in these categories are considered as white-box methods and they are not suitable in cases where a model's internal parameters are inaccessible.\\
\noindent
 \textbf{Black-box}: Contrary to the white-box methods, black-box methods do not require access to the model's internal parameters. For instance, all approaches in the category of surrogate methods (Sec. \ref{sec::surrogate_methods}) generate a local dataset using the model's input and output. Since these methods do not require access to the model parameters, all of them can be categorized as black-box methods.

\subsubsection{Decomposition-based Methods}
\label{sec:white_decom}

These methods consider the prediction of the model as a score that is decomposed and distributed backwards in a layer by layer fashion till it reaches the input. The score of different parts of the input can be construed as its importance to the prediction. However, the decomposition technique can vary across methods. They also require internal parameters of the model to calculate the score. Hence, these explanation methods are considered as white-box methods. Table \ref{tab::decomposition} provides a summary of these methods. 

\begin{table}[tb]
\vspace{-2mm}
  \centering
  \scriptsize
  \caption{Key highlights of \textit{decomposition-based} methods }
    \begin{tabular}{cccc}
    \toprule
          \textbf{Method} & \textbf{Parameters} & \textbf{Form of Explanation} & \textbf{Task}  \\  \midrule
        CAM \cite{Excitation-BP} & \makecell{Node embedding of last layer and MLP weights} & Node importance  & Graph Classification  \\  \hline
        Excitation-BP \cite{Excitation-BP} & \makecell{Weights of all GNN layers} &  Node importance & \makecell{Node and Graph Classification }  \\ \hline
        DEGREE \cite{degree} & \makecell{Decomposes messages and\\ requires all parameters} & Similar nodes & \makecell{Node and Graph Classification}  \\ \hline
        GNN-LRP \cite{GNN-LRP} & \makecell{Weights of all layers} & \makecell{Collection of edges} & \makecell{Node and Graph Classification} \\ 
        \bottomrule
    \end{tabular}%
  \label{tab::decomposition}%
\end{table}%

One of the decomposition-based methods, \textbf{CAM} \cite{Excitation-BP} aims at constructing the explanation of GNNs that have a Global Average Pooling (GAP) layer and a fully connected layer as the final classifier. Let $e_n$ be the final embedding of node $n$ just before the GAP layer and $w^c$ be the weight vector of the classifier for the class $C$. The importance score of the node $n$ is computed as $(w^c)^Te_n$. This means that the node's contribution to the class score $y^c$ is taken as the importance. It is clear that this method is restricted to GNNs that have a GAP layer and perform only the graph classification task.

Another method, \textbf{Excitation-BP} \cite{Excitation-BP} considers that the final probability of the prediction can be decomposed into excitations from different neurons. The output of a neuron can be intuitively understood as the weighted sum of excitations from the connected neurons in the previous layer combined with a non-linear function where the weights are the usual neural network parameters. With this, the output probability can be distributed to the neurons in the previous layer according to  the ratios of these weights. Finally, the importance of a node is obtained by combining the excitations of all the feature maps of that node.

Contrary to other methods, \textbf{DEGREE} \cite{degree} finds explanation in the form of subgraph structures. First, it decomposes the message passing feed-forward propagation mechanism of the GNN to find a contribution score of a group of target nodes. Next, it uses an agglomeration algorithm that greedily finds the most influential subgraph as the explanation. \textbf{GNN-LRP} \cite{GNN-LRP} is based on the concept that the function modeled by GNN is a polynomial function in the vicinity of a specific input. The prediction score is decomposed by approximating the higher order Taylor expansion using layer-wise relevance propagation \cite{LRP-main}. This differs from other decomposition-based methods not only in the decomposition technique but also in the score attribution. While other methods attribute scores to nodes or edges, GNN-LRP attributes scores to walks i.e., a collection of edges.

\subsubsection{Gradient-based Methods}
\label{sec::gradient-based}
The gradient-based explainer methods follow the following key idea. Gradients represent the rate of change, and the gradient of the prediction with respect to the input represents how sensitive the prediction is to the input. This sensitivity is seen as a measure of importance. We provide a summary of these methods in Table \ref{tab::wb-gradient}.

\textbf{Sensitivity Analysis (SA)} \cite{guided-bp} is one of the earlier methods to use gradients to explain GNNs. Let $x$ be an input, which can be a node or an edge feature vector, $SA(x)$ be its importance, and $\phi$ be the GNN model, then the importance is computed as $SA(x) \propto ||\nabla_{x} \phi(x)||^{2}$. The intuition behind this method is based on the aforementioned sensitivity to the input. \textbf{Guided Backpropagation (Guided-BP)} \cite{guided-bp}, a slightly modified version of the previous method SA, follows a similar idea except the fact that the negative gradients are clipped to zero during the backpropagation. This is done to preserve only inputs that have an excitatory effect on the output. Intuitively, since positive and negative gradients have opposing effect on the output, using both of them could result in less accurate explanations.


The method \textbf{Grad-CAM} \cite{Excitation-BP} builds upon \textbf{CAM} \cite{Excitation-BP} (see Section \ref{sec:white_decom}), and uses gradients with respect to the final node embeddings to compute the importance scores.  The importance score is $(\frac{1}{N}\sum_{n=1}^{N}\nabla_{e_n}(y^c))^Te_n$, where $e_n$ is the final embedding of node $n$ just before the GAP layer, and $w^c$ is the weight vector of the classifier for class $C$, $g = \frac{1}{N} \sum_{n=1}^{N} e_n$ is the vector after the GAP layer, and the final class score $y^c$ of \textit{C} is $w^Tg$.
This removes the restriction about the necessity of GAP layer. This equation shows that the importance of each node is computed as the weighted sum of the feature maps of the node embeddings, where the weights are the gradients of the output with respect to the feature maps.

All the above methods depend on this particular intuition that the gradients can be good indicators of importance. However, this might not be useful in many settings. Gradients indicates sensitivity which does not reflect importance accurately. Moreover, saturation regions where the prediction of the model does not change significantly with the input, can be seen as another issue in SA and Guided-BP.

\begin{table}[tb]
  \centering
  \vspace{-2mm}
  \scriptsize
  \caption{Key highlights of \textit{gradient-based} methods}
  \resizebox{\columnwidth}{!}{%
    \begin{tabular}{ccccc}
    \toprule
        \textbf{Method} & \textbf{Explanation Type} & \textbf{Task} & \textbf{Explanation Target} & \textbf{Datasets Evaluated}  \\  \midrule
        SA \cite{guided-bp} & Instance level & \makecell{Graph classification\\Node classification} & \makecell{Nodes, Node features\\Edges, Edge features} & \makecell{Infection, ESOL \cite{esol}} \\  \hline
        Guided-BP \cite{guided-bp} & Instance level & \makecell{Graph classification\\Node classification} & \makecell{Nodes, Node features\\Edges, Edge features} & \makecell{Infection, ESOL \cite{esol}} \\  \hline
        Grad-CAM \cite{Excitation-BP} & Instance level & \makecell{Node classification} & \makecell{Nodes, Node features} & \makecell{BBBP, BACE, TOX21 \cite{kersting2016benchmark}} \\ 
    \bottomrule
    \end{tabular}}
  \label{tab::wb-gradient}%
\end{table}%

\subsubsection{Surrogate Methods}
\label{sec::surrogate_methods}
\begin{figure}[htbp]
\vspace{-2mm}
     \centering
     \begin{subfigure}[b]{0.43\textwidth}
         \centering
         \includegraphics[width=\textwidth]{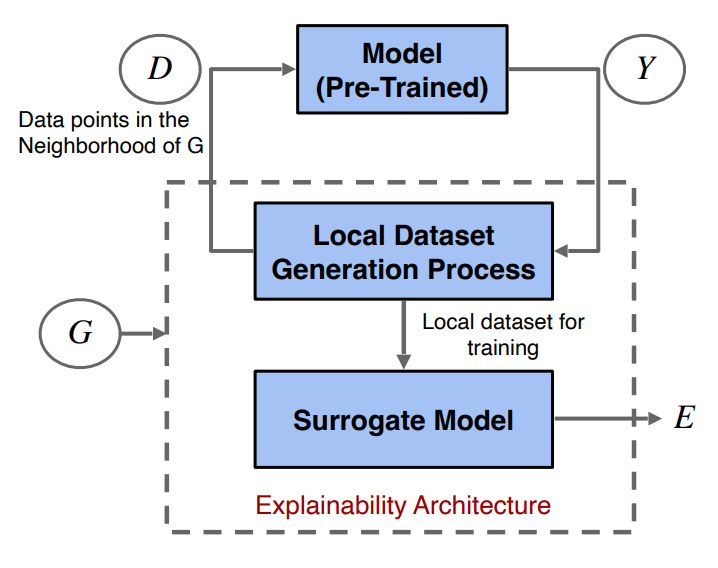}
         \caption{\textbf{Generic schema of surrogate methods}}
         \label{fig:Surrogate_schema}
     \end{subfigure}
     \hfill
     \begin{subfigure}[b]{0.54\textwidth}
         \centering
         \includegraphics[width=\textwidth]{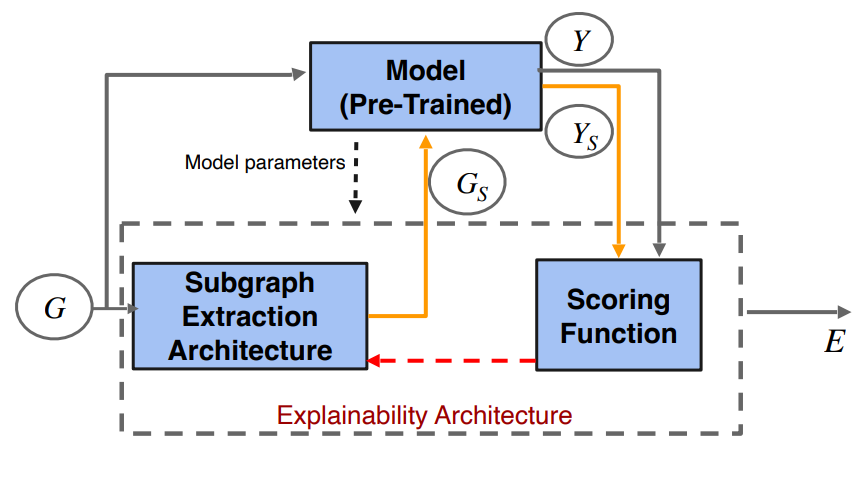}
         \caption{\textbf{Generic schema of perturbation-based methods}}
         \label{fig:Perturbation_schema}
     \end{subfigure}
        \caption{\small \textbf{a) Surrogate:} These methods follow a two-step process. For any instance \(\CG\), they generate data from the neighbourhood of the prediction by using multiple inputs \(D\) in the locality and recording its model prediction \(Y\). Then a surrogate model is used to fit this data. Explanation \(E\) for the surrogate model is the explanation for the prediction, \textbf{b) Perturbation-based:} They have two key modules: a subgraph extraction architecture and a scoring function. For an input \(G\), the subgraph extraction module extracts a subgraph \(G_s\). The model prediction \(Y_s\) for subgraph \(G_s\) are scored against the actual predictions \(Y\) using a scoring function. The feedback from the scoring function can be used to train the subgraph extraction module. Sometimes model parameters are also used as the training input to the subgraph extraction module. The optimal subgraph \(G_s^*\) acts as the final explanation \(E\). }
        \label{fig:surr_per}
\end{figure}

 Within a large range of input values, the relationship between input and output can be complex. Hence, we need complex functions to model this relationship and the corresponding model might not be interpretable. However, in a smaller range of input values, the relationship between input and output can be approximated by simpler and interpretable functions. This intuition leads to \textit{surrogate methods} that fit a simple and interpretable surrogate model in the locality of the prediction. Table \ref{tab::surrogate} shows different locality-based data extraction techniques and surrogate models used by surrogate methods. This surrogate model can then be used to generate explanations. As seen in the figure \ref{fig:Surrogate_schema}, these methods adopt a two-step approach. Given an instance \(\CG\), they first generate data from the prediction's neighborhood by utilizing multiple inputs  \(D\) within the vicinity and recording the model's prediction \(Y\). Subsequently, a surrogate model is employed to train on this data. The explanation \(E\) provided by the surrogate model serves as an explanation for the original prediction.  

\textbf{PGMExplainer} constructs a Bayesian network to explain the prediction. First, it creates a tabular dataset by random perturbations on node features of multiple nodes of the computational graph and records its influence on the prediction. A grow-shrink algorithm is used to select top influential nodes. Using structure learning, a Bayesian network is learnt that optimizes the Bayesian Information Criterion (BIC) scores and the DAG of conditional probabilities act as an explanation. In \textbf{GraphLime} \cite{graphlime}, the local explanations are based on Hilbert-Schmidt Independence Criterion Lasso (HSIC Lasso) model, which is a kernel-based nonlinear interpretable feature selection algorithm. This method assumes that the node features in the original graph are easily interpretable. The HSIC model takes a node and its N-hop neighbourhood (for some $N$), and selects a subset of node features that are the most influential to the prediction. These selected features act as the explanation. To construct a local dataset, \textbf{GraphSVX} \cite{graphsvx} uses a mask generator to jointly perturb the nodes and the features and observes its effects on the predictions. The mask generator isolates the masked nodes and replaces masked features by its expected values. It then fits a weighted linear regression model (WLR) on the local dataset. The coefficients of WLR act as explanations.

The next two approaches use GNN based models to act as surrogate models. \textbf{RelEx} \cite{RELex} uses a BFS-based sampling strategy to select nodes and then perturb them to create the local dataset. Then, a GCN model with residual connections is used to fit this dataset. In contrast to other methods in this category, the surrogate model of RelEx is not interpretable. Hence, it uses perturbation-based strategy to find a mask that acts as explanation. We note that the surrogate model is more complex compared to other methods and it requires the use of another explanation method to derive explanations from the surrogate model. \textbf{DistilnExplain (DnX)}~\cite{distilexplain} first learns a surrogate GNN via knowledge distillation and then provides an explanation by solving a simple convex program. In contrast to RelEx, DnX uses a simpler surrogate model which is a linear architecture termed as Simplified Graph convolution (SGC) \cite{Simple-GCN}. SGC does not have any non-linear activation layers and uses a single parameter matrix across layers. The parameters in SGC are learned via knowledge distillation with an objective to minimize the KL-divergence between the predictions by SGC and the model. Furthermore, explanations can be derived from SGC by solving a simple convex program.

\begin{table}[tb]
  \centering
  \scriptsize
  \caption{Key highlights of \textit{surrogate methods} }
    \begin{tabular}{ccccc}
    \toprule
          \textbf{Method} & \textbf{Local Dataset extraction} & \textbf{Surrogate model} & \textbf{Explanation} & \\  \midrule
        GraphLime \cite{graphlime} & N-hop neighbor nodes & HSIC Lasso  & Weights of the model & \\  \hline
        PGMExplainer \cite{pgm-ex} & \makecell{Random node \\feature perturbation}  & Bayesian network  & DAG of conditional dependence & \\  \hline
        ReLex \cite{RELex} & \makecell{Random sampling \\of connected subgraphs} & GCN & Perturbation-based method  &\\ \hline
        DistilnExplain \cite{distilexplain} & Entire dataset  & Knowledge distilled Simple GCN \cite{Simple-GCN}  & \makecell{Convex programming,\\ decomposition} &  \\ \hline
        GraphSVX \cite{graphsvx} & \makecell{Input perturbations \\via mask generator}  & Weighted Linear Regression (WLR)   & Weights of WLR &  \\ \bottomrule
        
    \end{tabular}%
  \label{tab::surrogate}%
\end{table}%

 \subsubsection{Perturbation-based Methods} 
 \label{sec::perturbation}
 These methods find \textit{important subgraphs} as explanations by perturbing the input. Fig. \ref{fig:Perturbation_schema} presents two key modules of these methods: the subgraph extraction module and the scoring function module. For an input \(G\), the subgraph extraction module extracts a subgraph \(G_s\). The model predictions \(Y_s\) for subgraphs are scored against the actual predictions \(Y\) using a scoring function. The feedback from the scoring function can be used to train the subgraph extraction module. In some cases, model parameters are also used as the training input for the subgraph extraction module. These methods provide explanations \(E\) in the form of a subgraph structure and some also provide node features as explanations. Table \ref{tab::perturbation} presents a summary of these methods.


\textbf{GNNExplainer} \cite{ying2019gnnexplainer} is one of the initial efforts towards the explainability of GNNs. It identifies an explanation in the form of a Subgraph including a subset of node features that have the maximum influence on the prediction. It learns continuous masks for both adjacency matrix and features by optimizing cross entropy between the class label and model prediction on the masked subgraph. In
 a follow-up work, \textbf{PGExplainer} \cite{pgexplainer} extends the idea in GNNExplainer by assuming the graph to be a random Gilbert graph, where the probability distribution of edges is conditionally independent. The distribution of each edge is independently modeled as a Bernoulli distribution, i.e., each edge has a different parametric distribution. These parameters are modeled by a neural network (MLP), and the parameters of this MLP is computed by optimizing the mutual information between the explanation subgraph and the predictions of the underlying GNN model. Another masking-related method, \textbf{GraphMask} \cite{Graph-mask} provides an explanation by learning a parameterized edge mask that predicts the edge to drop at every layer. A single-layer MLP classifier is trained to predict the edges that can be dropped. To keep the topology unaffected, these edges are not dropped but are replaced by a learned baseline vector. The training objective is to minimize the $L_0$ norm i.e., the total number of edges not masked, such that the prediction output remains within a tolerance level. To make the objective differentiable, it uses sparse relaxations through the reparameterization trick and the hard concrete distribution \cite{concrete-distri, repara-trick}. 
 
 Another approach \textbf{Zorro} \cite{zorro} finds explanations in the form of important nodes and features that maximizes \textit{Fidelity} (see Sec. \ref{sec::eval}). It uses a greedy approach that selects the node and the feature at each step with the highest fidelity score. Fidelity is computed as the expected validity of the perturbed input. The approach uses a discrete mask for selecting a subgraph without any backpropagation.
A two-staged approach, \textbf{ReFine} \cite{ReFine} consists of the edge attribution or pre-training and the edge selection or fine-tuning steps. During pre-training, a GNN and an MLP are trained to find the edge probabilities for the entire class by maximizing mutual information and contrastive loss between classes. During the fine-tuning step, the edge probabilities from the previous stage are used to sample edges and find an explanation that maximizes mutual information for a specific instance.

The next two approaches use cooperative game theoretic techniques. \textbf{SubgraphX} \cite{subgraphX} applies the Monte Carlo Tree search technique for subgraph exploration and uses the Shapley value \cite{Shap_val} to measure the importance of the subgraphs. For the search algorithm, the child nodes are obtained by pruning the parent graph. In computing the Shapley values, the Monte Carlo sampling helps to find a coalition set and the prediction from the GNN is used as the pay-off in the game. In a subsequent work, \textbf{GStarX} \cite{gstarx} uses a different technique from cooperative game theory known as HN value \cite{HN-val}, to compute importance scores of a node for both graph and node classification tasks. In contrast to the Shapley value, the HN value is a structure-aware metric. Since computing the HN values is expensive, Monte Carlo sampling is used for large graphs. The nodes with the top-k highest HN values act as an explanation.

\begin{table}[tb]
\scriptsize
  \centering
  \caption{Key highlights of the \textit{perturbation-based} methods. Note that MI is mutula information, SV is Shapley value, and Explanation denotes node feature explanation.}
    \begin{tabular}{ccccc}
    \toprule
          \textbf{Method} & \textbf{Subgraph Extraction Strategy} & \textbf{Scoring function} & \textbf{Constraints} & \textbf{Explanation}  \\  \midrule
        GNNExplainer \cite{ying2019gnnexplainer} & Continuous relaxation  & MI  & Size  & Yes  \\  
        SubgraphX \cite{subgraphX} & Monte Carlo Tree Search & SV  & Size, connectivity  & No \\  
        GraphMask \cite{Graph-mask} & Layer-wise parameterized edge selection &  $L_0$ norm  & Prediction divergence & No \\  
        PGexplainer \cite{pgexplainer} & Parameterized edge selection & MI & Size and/or connectivity & No \\  
        Zorro \cite{zorro} & Greedy selection & Fidelity & Threshold fidelity & Yes  \\  
        ReFine \cite{ReFine} & Parameterized edge attribution & MI  & Number of edges  & No \\  
         GstarX \cite{gstarx} & Monte Carlo sampling & HN-value  & Size  & No  \\  \bottomrule
    \end{tabular}%
  \label{tab::perturbation}%
\end{table}%

\begin{table}[tb]
  \centering
  \scriptsize
  \caption{Key highlights of the \textit{generation-based} methods }
    \begin{tabular}{ccccc}
    \toprule
          \textbf{Methods} & \textbf{Explanation Type} & \textbf{Optimization} & \textbf{Constraints} & \textbf{Task}  \\  \midrule
        XGNN \cite{xgnn} & Model level & RL-policy gradient  & Domain specific rules  & Graph classification  \\  
        RG-Explainer \cite{RL-enhanced} & Instance level & RL-policy gradient  & Size, radius, similarity & Node \& Graph classification \\  
        GFLOW Explainer \cite{Gflow} & Instance level & TD flow matching  & Connectivity / cut vertex  & Node \& Graph classification \\  
        GNNinterpreter \cite{gnninterpreter} & Model level & Continuous relaxation  & Similarity to mean & Graph Classification \\  
        GEM \cite{Gen-causal}& Instance level & Autoencoder  & Graph validity rules  & Node \& Graph Classification \\  
        \bottomrule
    \end{tabular}%
  \label{tab::generation}%
\end{table}%

\subsubsection{Generation-based methods}
\label{sec::generation-based}
Generation-based approaches either use generative models or graph generators to derive instance-level or model-level explanations.
Furthermore, to ensure the validity of the generated graphs, different approaches have been proposed. Table \ref{tab::generation} provides a summary of the generation-based methods.    

\textbf{XGNN}~\cite{xgnn} provides model-level explanations by generating key subgraph patterns to maximize prediction for a certain class. The subgraph is generated using a Reinforcement learning (RL) based graph generator which is optimized using policy gradient. In the setup for the RL agent, the previous graph is the state; adding an edge is an action; and the model prediction along with the validity rules acts as the reward. Unsurprisingly, the validity rules are specified based on domain knowledge.  Another RL-based method, \textbf{RG-Explainer} \cite{RL-enhanced} formulates the underlying problem as combinatorial optimization instead of using continuous relaxation or search methods to find the subgraph. A starting point is selected using an MLP which acts as an input to the graph generator. The graph generator is an RL agent that optimizes for the policy using policy gradient with subgraph as the state, adding neighboring nodes as the action, and the function of the cross entropy loss as the reward. 

A non-RL method, \textbf{GNNinterpreter} \cite{gnninterpreter} is a generative model-level explanation method for the graph classification task. Its objective is to maximize the likelihood of predicting the explanation graph correctly for a given class. The similarity between the explanation graph embedding and the mean embedding of all graphs act as an optimization constraint. Intuitively, this ensures that the explanation graph stays closer to the domain and is meaningful. Since the adjacency matrix and sometimes even the features can be categorical, GNNinterpreter uses the Grumbel softmax method \cite{repara-trick} to enable backpropagation of gradients. Contrary to XGNN with domain-specific hand-crafted rules, GNNinterpreter uses numerical optimization and does not need any domain knowledge.

\textbf{GFLOW Explainer} \cite{Gflow} uses GFLOWNETs as the generative component. The objective is to construct a TD-like flow matching condition \cite{bengio-gflownet} to learn a policy to generate a subgraph by sequentially adding neighbors (nodes) such that the probability of the subgraph of a class is proportional to the mutual information between the label and the distribution of possible subgraphs. A \textit{state} consists of several nodes with the initial state as the single most influential node and the end state that satisfies the stopping criteria. \textit{Action} is adding a node and the \textit{reward} is a function of the cross-entropy loss.

 \textbf{GEM~\cite{Gen-causal}} uses the principles of Granger causality to generate ground-truth explanations which are used to train the explainer. It quantifies the causal contribution of each edge in the computational graph by the difference in the loss of the model with and without the edge. This distilled ground-truth for the computation graph is used to train the generative auto-encoder based explainer. This explainer provides an explanation for any instance in the form of the subgraph of the computation graph.


 


\subsection{Self-interpretable} 
\begin{figure*}[t]
  \centering
  \vspace{-5mm}
  \includegraphics[width= .8\textwidth]{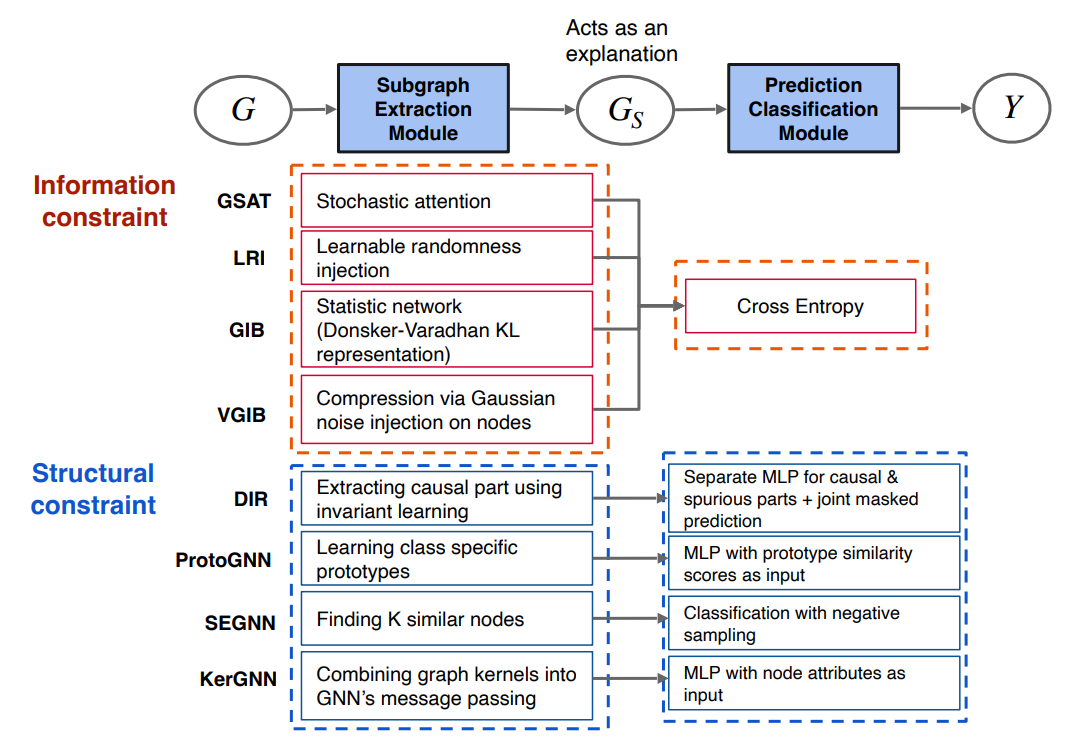}
  \centering
  \vspace{-1mm}
  \caption{\small \textbf{Self-interpretable methods: } Every self-interpretable method has a \textit{subgraph extraction} and a \textit{prediction} module. The subgraph extraction module (the function \(g\)) uses constraints to find an informative subgraph \(\CG_s\) from input graph \(\CG\). The prediction module uses \(\CG_s\) to predict label \(Y\). This also shows the techniques used by each method to implement these individual modules. Self-interpretable Methods are categorized based on constraints: \textbf{(1) Information constraint:} GIB \cite{GIB}, VGIB \cite{VGIB}, GSAT \cite{GSAT}, LRI \cite{inject-explain}; \textbf{(2) Structural constraint:} DIR \cite{D_invariant_rationale}, ProtGNN \cite{protgnn}, SEGNN \cite{SE-GNN}, KER-GNN \cite{kergnns}.}

  \label{fig:SE_summary}
\end{figure*}

In self-interpretable methods, the explainable procedure is intrinsic to the model. Such methods derive explainability by incorporating interpretability constraints. These methods use either information constraints or cardinality (structural) constraints to derive an informative subgraph which is used for both the prediction and the explanation. Based on the design of the explainability, we further classify the self-interpretable methods into two types based on the imposed constraints (Fig. \ref{fig:SE_summary}). 

\subsubsection{Methods with information constraints }
\label{sec:information_const}

One of the major challenges in constructing explanations via subgraphs is that the critical subgraphs may have different sizes and can be irregular. Thus, constraining the size of the explanation may not be appropriate for the underlying prediction task. To address this challenge, the methods based on information constraint use the principle of information bottleneck (IB) \cite{ib_principle} to impose constraints on the information instead of the size. For a graph \(\CG\), subgraph \(\CG_s\) and label \(Y\), the graph information bottleneck (GIB) objective is: 
\begin{equation*}
    \max_{\CG_s} I(Y,\CG_s) \; \text{  such that  } \: I(\CG,\CG_s) \leq  \gamma
\end{equation*}
where \(I\) denotes the mutual information. Using Lagrangian multiplier $\beta$, we can write the equation as:
\begin{equation*}
    \min_{\CG_s} - I(Y,\CG_s) \ + \: \beta* I(\CG,\CG_s)
\end{equation*} 


As seen from the equations, GIB objective-based methods have two parts in the objective function and both are intractable. All methods approximate \(I(Y,\CG_s)\) by calculating the cross-entropy loss. However, all methods vary in their approach in making \(I(\CG,\CG_s)\) tractable i.e., all have different approaches to compressing the graph and finding the informative subgraph \(\CG_s\). This subgraph is used for both prediction and interpretation. Table \ref{tab::Info_constraint} provides the summary of all methods in this category.
\begin{table}[tb]
  \centering
  \scriptsize
  \caption{Key highlights of the methods with \textit{information constraints} }
    
    \begin{tabular}{cccc}
    \toprule
          \textbf{Method} & \textbf{Process of calculating \(I(\CG,\CG_s)\) } & \textbf{Injection of Randomness} & \textbf{Subgraph Extractor Architecture}  \\  \midrule
        GSAT \cite{GSAT} & Stochastic attention & Bernoulli as Prior for KL divergence & GNN + MLP + Reparameterization \\  \hline
        LRI \cite{inject-explain} & Learnable Randomness injection  & Bernoulli and Gaussian as prior & GNN + MLP + Reparameterization  \\ \hline
        GIB \cite{GIB} & Donsker-Vardhan KL representation \cite{KL_donsker}  & No randomness injection & Statistic Network: GNN + MLP \\ \hline
        VGIB \cite{VGIB} & Compression via Noise injection  & Gaussian noise on node features & GNN + MLP + Reparameterization \\ \bottomrule

    \end{tabular}%
    
  \label{tab::Info_constraint}%
\end{table}%

\textbf{GSAT} \cite{GSAT} uses a stochastic attention mechanism to calculate the variational upper bound for \( I(\CG,\CG_s)\). First, it encodes graph \(\CG\) using a GNN to find the representation for each node. Then, for each node pair \((u,v)\), GSAT uses an  MLP to calculate \(P_{uv}\). This is used to sample stochastic attention from Bernoulli distribution \(Bern(P_{uv})\) to extract a subgraph \(\CG_s\). The variational upper bound is the KL divergence between \(Bern(P_{uv})\) and \(Bern(\alpha)\) where \(\alpha\) is a hyper-parameter. Building on similar concepts, \textbf{LRI} \cite{inject-explain} uses both Bernoulli and Gaussian distribution as the prior distribution. LRI-Bernoulli provides the existence importance of points and LRI-Gaussian provides the location importance of the points i.e., how perturbing the location of the point in different directions affects the prediction. Another method, \textbf{GIB} \cite{GIB} assumes that there is no reasonable prior distribution to solve \( I(\CG,\CG_s)\) via KL divergence in the graph space. Hence, it uses the Donsker-Vardhan KL representation \cite{KL_donsker} in the latent space. It employs a bi-level optimization wherein the statistic network of the Donsker-Varadhan representation is used to estimate \( I(\CG,\CG_s)\) in the inner loop. This estimate with classification and connectivity loss is used to optimize the GIB objective in the outer loop. This bi-level training process is inefficient and unstable; and hence \textbf{VGIB} \cite{VGIB} uses a different compression technique. The information in the original graph is dampened by injecting noise into the node representations via a learned probability \(P_i\) for each node \(i\). The classification loss will be higher if the informative substructure \(\CG_s ^ *\) is injected with noise. Hence, \(\CG_s ^*\) is less likely to be injected with noise compared to label-irrelevant substructures.

\subsubsection{Methods with structural constraints}
\label{subsec:: SE- structural-constr}
Imposing structural constraints on the input to derive the most informative subgraph has also been a common approach. The obtained informative subgraph is used for both making predictions and generating explanations. The key difference across the methods is the set up of the structural constraints. In Table \ref{tab::struct_constraint}, we provide the key highlights of these methods.

\begin{table}[tb]
  \centering
  \scriptsize
  \caption{Key highlights of explainability methods with \textit{structural constraints}. Note that NC and GC denote node and graph classification respectively.}
    \begin{tabular}{ccccc}
    \toprule
          \textbf{Method} & \textbf{Subgraph Extraction} & \textbf{Explanation form} &\textbf{Prediction/ classification module} & \textbf{Task}  \\  \midrule
        DIR \cite{D_invariant_rationale} & \makecell{Separating pattern that is invariant \\ across interventional distribution} & Invariant rationale  & \makecell{Seperate MLP for spurious\\ and invariant parts }   & GC \\  \hline
        ProtoGNN \cite{protgnn} & \makecell{Computes similarity between \\Graph embedding and several \\learned diverse prototypes} &  \makecell{Prototypes with \\ high similarity} & \makecell{MLP with similarity \\ scores as input }& GC  \\ \hline
        SEGNN \cite{SE-GNN} & \makecell{Finds K nodes that have similar\\ structure and node features\\ via contrastive loss} & Similar nodes & \makecell{ classification via negative\\ sampling of nodes}  & NC \\ \hline
        KER-GNN \cite{kergnns} & \makecell{Kernel filters integrated  in message \\passing of GNNs} & \makecell{learned kernels and \\ output node attributes} & \makecell{MLP with node \\attributes as input} & \makecell{NC,\\ GC} \\ \bottomrule

    \end{tabular}%
  \label{tab::struct_constraint}%
\end{table}%

One of the earlier methods, \textbf{DIR} \cite{D_invariant_rationale} finds explanations in the form of invariant causal rationales by learning to split the input into causal ($C$) and non-causal ($S$) parts. The objective is to minimize the classification loss such that $Y$ (the prediction) is independent of $S$ given $C$. To achieve this, it first creates multiple interventional distributions by conducting interventions on the training distribution. The part that is invariant across these distributions is considered a causal part. Moreover, the implementation has three key stages. First, the architecture consists of a rationale generator (GNN) that splits the input graph into a causal part with top $k$ edges and a non-causal part. Second, a distribution intervener, i.e., a random replacement from a set, creates perturbed distribution to infer the invariant causal parts. Finally, two classifiers are used to generate a joint prediction on causal and non-causal parts.


\textbf{ProtoGNN} \cite{protgnn} combines prototype learning \cite{prototype-learning} with GNNs. Prototype learning is a form of case-based reasoning which makes predictions for new instances by comparing them with several learned exemplar cases also called prototypes. ProtoGNN computes the similarity scores between the graph embedding and multiple learned prototypes. Moreover, these prototypes are projected onto the nearest latent training subgraph during training using the Monte Carlo tree search \cite{monte-carlo,monte-carlo-survey}. The similarity scores are used for the classification task where the subgraphs with high similarities can be used for explanation. In another work, for a given unlabeled node, \textbf{SEGNN} \cite{SE-GNN} finds $k$ nearest labeled nodes that have structural and feature similarities and can be used for both generating predictions and explanations. It uses contrastive loss on node representations for feature similarity and also on edge representations of local neighborhood nodes for structural similarity. Moreover, the classification loss uses negative sampling with approximate $k$ similar nodes. These $k$ nearest nodes can be used to derive an explanation subgraph with threshold importance.

The method, \textbf{KER-GNN} \cite{kergnns} integrates graph kernels into the message-passing process of GNNs to increase the expressivity of GNNs beyond the 1-WL isomorphism test. In each layer, the node embeddings are updated by computing the similarity between the node's subgraph (the node with its ego-net) and trainable filters in the form of hidden graphs. The learned graph filters can provide important structural information about the data. Moreover, the output node attributes can be used to extract important substructures.

\section{Counterfactual Explanation}
\label{sec::Counterfactual}
Counterfactual methods provides an explanation by identifying the minimal alteration in the input graph that results in a change in the model's prediction. Recently, there have been several attempts to have explanations of graph neural networks (GNNs) via counterfactual reasoning. We classify these explainer methods that find counterfactuals into three major categories based on the type of methods: \textbf{(1) Perturbation-based}, \textbf{(2) Neural framework-based}, and \textbf{(3) Search-based}. We discuss the works in the individual categories below. 

\subsection{Perturbation-based methods} 
An intuitive way to generate counterfactuals for both the graph classification and the node classification task is to \textit{alter the edges}, i.e., add or delete the edges in the graph such that it would change the prediction of the underlying GNN method. This alteration can be achieved by perturbing either the adjacency matrix or the computational graph of a node. The perturbation-based methods are summarized in Table \ref{tab::cf-perturbation}.

One of the initial efforts, \textbf{CF-GNNExplainer} \cite{cfgnnex} aims to perturb the computational graph by using a binary mask matrix. It uses a binary matrix (all values are 0 or 1) $P$ and modifies the computational graph matrix as $\Tilde{A_v} = P \odot A_v$, where $A_v$ is the original computational graph matrix and $\Tilde{A_v}$ is computational graph matrix after the perturbation. The matrix $P$ is computed by minimizing a combination of two different loss functions: $L_{pred}$, and $L_{dist}$. They are combined using a hyper-parameter in the final loss ($L$) as $L_{pred} + \beta L_{dist}$. The loss function, $L_{pred}$ quantifies the accuracy of the produced counterfactual, and $L_{dist}$ captures the distance (or similarity) between the counterfactual graph and the original graph. In follow-up work, the method \textbf{CF\textsuperscript{2}} \cite{cf^2-counter} extends the method in CF-GNNExplainer \cite{cfgnnex} by including a contrastive loss that jointly optimizes the quality of both the factual explanation and the counterfactual one. For an input graph, $G$, it aims to find an optimal subgraph $G_s$ where $G_s$ is a good factual explanation, and $G \backslash G_s$ is a good counterfactual. These objectives are formulated as a single optimization problem with the corresponding loss as $L_{overall} = \alpha L_{factual} + (1 - \alpha)L_{counterfactual}$, where $\alpha$ is a hyperparameter.

Another method, \textbf{GREASE}~\cite{chen2022grease} follows the standard technique of using a perturbation matrix to generate a counterfactual, but with two key modifications mainly to accommodate GNNs used for recommendation systems instead of classification tasks. In the recommendation task, GNNs rank the items (nodes) by assigning them a score instead of classifying them. GREASE uses a loss function based on the scores given by the GNN before and after the perturbation. This score helps to rank the items or nodes. The second modification is the perturbation matrix, which acts as the mask, and is used to perturb the computational graph (\textit{l}-hop neighborhood of the node) instead of perturbing the entire graph. Here \textit{l} denotes the number of layers in the GNN. Similar to CF\textsuperscript{2} \cite{cf^2-counter}, GREASE also optimizes counterfactual and factual explanation losses, but not jointly. 


In summary, all these techniques share similarities in computing the counterfactual similarity and constructing the search space. Similarity is measured by the number of edges removed from input instances and the search space is the set of all subgraphs obtained by edge deletions in the original graph. Because of the unrestricted nature of the search space, these methods might not be ideal for graphs such as molecules, where the validity of the subgraphs has valency restrictions. On the other hand, the mentioned methods differ mainly in the loss function formulations and the perturbation operations for the downstream tasks. For instance, CF-GNNExplainer \cite{cfgnnex} and GREASE \cite{chen2022grease} perform node classification and regression, they can use perturbations on the computation graph. However, CF\textsuperscript{2} \cite{cf^2-counter} considers both graph and node classification tasks, hence it uses perturbations on the entire graph, i.e., the adjacency matrix.

\begin{table}[tb]
  \centering
  \scriptsize
  \caption{Key highlights of \textit{perturbation-based} methods for counterfactuals}
    \begin{tabular}{ccccc}
    \toprule
        \textbf{Method} & \textbf{Explanation Type} & \textbf{Downstream Task} & \textbf{Perturbation Target} & \textbf{Datasets Evaluated}  \\  \midrule
        CF-GNNExplainer \cite{cfgnnex} & Instance level  & Node Classification  & Computation graph  & \makecell{Tree-Cycles \cite{ying2019gnnexplainer}, Tree-Grids \cite{ying2019gnnexplainer}\\BA-Shapes \cite{ying2019gnnexplainer}}  \\  \hline
        CF\textsuperscript{2} \cite{cf^2-counter}  &  Instance level & \makecell{Graph Classification\\Node Classification}  & Original graph & \makecell{BA-Shapes \cite{ying2019gnnexplainer}, Tree-Cycles \cite{ying2019gnnexplainer} \\ Mutag \cite{mutag}, NCI1 \cite{NCI1_data}, CiteSeer \cite{getoor2005advanced}} \\  \hline
        GREASE \cite{chen2022grease}  & Instance level  &  Node Ranking & Computation graph & LastFM, Yelp \\  \bottomrule
    \end{tabular}%
  \label{tab::cf-perturbation}%
\end{table}%

\subsection{Neural framework-based methods}
The approaches in this section use neural architectures to generate counterfactual graphs as opposed to the perturbation-based methods where the adjacency matrix of the input graph is minimally perturbed to generate counterfactuals. Table \ref{tab::cf-neural} summarizes these methods.

The objective of \textbf{RCExplainer} \cite{robust-counter} is to identify a resilient subset of edges that, when removed, alter the prediction of the remaining graph. This is accomplished by modeling the implicit decision regions using graph embeddings.
Even though the counterfactual graph generated by a neural architecture is used in conjunction with the adjacency matrix of the input graph, the counterfactual itself is not generated through perturbations on the adjacency matrix. RCExplainer addresses the issue of fragility where an interpretation is fragile (or non-robust) if systematic perturbations in the input graph can lead to dramatically different interpretations without changing the label. The standard explainers aim to generate good counterfactuals by choosing the closest counterfactual to the input instance and it might induce over-fitting. RCExplainer reduces this over-fitting by first clustering input graphs using polytopes, and finding good counterfactuals close to the cluster (polytope) instead of individual instances. Another method, \textbf{CLEAR} \cite{clear-counter} generates counterfactual graphs by leveraging a graph variational autoencoder. Two major issues often seen in other explainer methods, namely, generalization and causality are addressed in this paper.

Both methods use a generative neural model to find counterfactuals, but the generative model is different across the methods. While \textbf{RCExplainer} \cite{robust-counter} uses a neural network that takes pairwise node embeddings and predict the existence of an edge between them, \textbf{CLEAR} \cite{clear-counter} uses a variational autoencoder to generate a complete graph. This shows that while the former method cannot create nodes that are not present in the original graph, the latter can. In terms of the objective, the primary focus in \textbf{RCExplainer} \cite{robust-counter} is the robustness of the generated counterfactual, but \textbf{CLEAR} \cite{clear-counter} aims to generate counterfactuals that explain the underlying causality.


\begin{table}[tb]
\vspace{-2mm}
\centering
  \scriptsize
  \caption{Key highlights of \textit{neural framework-based} methods for counterfactuals}
    \begin{tabular}{ccccc}
    \toprule
        \textbf{Method} & \textbf{Explanation Type} & \textbf{Downstream Task} & \textbf{Counterfactual Generator} & \textbf{Datasets Evaluated}  \\  \midrule
        RCExplainer \cite{robust-counter} & Instance level & \makecell{Graph classification\\Node classification} & \makecell{Edge prediction\\with Neural Network} & \makecell{Mutag \cite{mutag}, BA-2motifs \cite{pgexplainer}, NCI1 \cite{NCI1_data}\\ Tree-Cycles \cite{ying2019gnnexplainer}, Tree-Grids \cite{ying2019gnnexplainer} \\BA-Shapes \cite{pgexplainer}, BA-Community \cite{ying2019gnnexplainer}} \\  \hline
        CLEAR \cite{clear-counter} & Instance level & \makecell{Graph classification\\Node classification} & \makecell{Graph generation\\with Variational Autoencoder} & \makecell{Community \cite{erd1959and}, Ogbg-molhiv,  IMDB-M} \\  \bottomrule
    \end{tabular}%
    \vspace{-1mm}  \label{tab::cf-neural}%
\end{table}%



\subsection{Search-based methods}
\label{sec::Counterfactual-search}
These methods usually depend on search techniques over the counterfactual space for relevant tasks or applications (see the highlights in Table \ref{tab::cf-search}). For example, given an inactive molecule in a chemical reaction, the task is to find a similar but active molecule. Here, generative methods or perturbation methods might not be effective, and the perturbations might not even result in a valid molecule. In such cases, a good search technique through the space of counterfactuals could be more useful. An inherent challenge is that the search space of counterfactuals might be exponential in size. Hence, building efficient search algorithms is required.

The major application is finding counterfactual examples for molecules in related tasks. The method \textbf{MMACE} \cite{agnostic-counter} finds counterfactuals for molecules.
In the corresponding graph classification problem, it aims to classify a molecule based on a specific property. Examples include whether a molecule will permeate blood brain barrier and molecule's solubility.
The search space can be generated by a method called \textit{Superfast Traversal, Optimization, Novelty, Exploration and Discovery (STONED)} \cite{nigam2021beyond}. MMACE uses this method to generate the close neighbourhood and searches with a BFS-style algorithm to find an optimal set of counterfactuals.

Similarly, \textbf{MEG} \cite{meg-counter} also aims to find a counterfactual and the search space consists of molecules. However, instead of searching the space with traditional graph search algorithms, MEG uses a reinforcement learning-based approach to navigate the search space more efficiently. The reward for finding a counterfactual is defined as the inverse of the probability that the candidate molecule found by the agent is not a counterfactual. This method is applied in a classification problem of predicting toxicity of a molecule as well as in a regression problem of predicting solubility of a molecule.

Another approach \textbf{GCFExplainer} \cite{Global-counter} uses a random walk-based method to search the counterfactual space. The objective is not to find an individual counterfactual for each input sample but to find a small set of counterfactuals that explain all or a subset of the input samples. Hence, this is a global method (see Sec. \ref{sec::global}). Here the counterfactual search space is obtained by applying graph edit operations on the training data. The method uses a  random walk called \textbf{Vertex Reinforced Random Walk (VRRW)} \cite{pemantle1992vertex}, which is a modified version of a Markov chain where the state transition probabilities depend on the number of previous visits to that state.



Both \textbf{MMACE} \cite{agnostic-counter} and \textbf{MEG} \cite{meg-counter} are developed for GNNs that predict molecular properties while the objective of \textbf{GCFExplainer} \cite{Global-counter} is to generate global explanations. However, the search algorithms and the generation mechanisms of the counterfactual space are quite different. For instance, MMACE employs a graph search algorithm to locate the nearest counterfactual instance. In contrast, MEG utilizes reinforcement learning, and GCFExplainer employs random walks to achieve the same.


\begin{table}[tb]
\vspace{-3mm}  
\centering
  \scriptsize
  \caption{Key highlights of \textit{search-based} methods for counterfactuals}
    \begin{tabular}{ccccc}
    \toprule
        \textbf{Method} & \textbf{Explanation Type} & \textbf{Downstream Task} & \textbf{Counterfactual Similarity Metric} & \textbf{Datasets Evaluated}  \\  \midrule
        MMACE \cite{agnostic-counter} & Instance level & \makecell{Graph classification\\Node classification} & Tanimoto similarity & \makecell{Blood brain barrier dataset \cite{martins2012bayesian}\\Solubility data \cite{sorkun2019aqsoldb}\\HIV drug dataset \cite{meg-counter} } \\  \hline
        GCFExplainer \cite{Global-counter} & Global & \makecell{Graph classification} & Graph edit distance & \makecell{Mutag \cite{mutag}, NCII \cite{NCI1_data} } \\  \hline
        MEG \cite{meg-counter} & Instance level & \makecell{Graph classification\\Node classification} & \makecell{Cosine Similarity\\Tanimoto similarity} & \makecell{Tox21 \cite{kersting2016benchmark}, ESOL \cite{moleculenet}} \\  \bottomrule
    \end{tabular}%
    \vspace{-3mm}
  \label{tab::cf-search}%
\end{table}%
\section{Others}

\label{sec::Others}
In this section we describe the explainer methods in three special categories: \textbf{explainer for temporal GNNs, global explainers} and \textbf{causality-based explainers} in this section.

\subsection{Explainers for Temporal GNNs}

In temporal or dynamic graphs, the graph topology and node attributes evolve over time. For instance, in social networks the relationships can be dynamic, or in the citation networks co-authorships change over time. There has been effort towards explaining the GNN models that are specifically designed for such structures. 

One of the earlier explainer methods on dynamic graphs is GCN-SE \cite{fan2021gcn}. GCN-SE learns the attention weights in the form of linear combination of the representations of the nodes over multiple snapshots (i.e., over time). To quantify the explanatory power of the proposed method, importance of different snapshots are evaluated via these learned weights. Another method \cite{he2022explainer} designs a two-step process. It uses static explainer such as PGM-explaine~\cite{pgm-ex} to explain the underlying temporal GNN (TGNN) model (such as TGCN~\cite{zhao2019tgcn}) for each time step separately and then it aims to discover the dominant explanations from the explanations identified by the static one. DGExplainer \cite{xie2022explaining} also generates
explanations for dynamic GNNs by computing the
relevance scores that capture the contributions of each component for a dynamic graph. More specifically, it redistributes the output activation score to the relevance of the neurons of its previous layer in the model. This process iterates until the relevance scores of the input neuron are obtained. Recently, T-GNNExplainer~\cite{xiaexplaining} has been proposed for temporal graph explanation where a temporal graph constituted by a sequence of temporal events. T-GNNExplainer solves the problem of finding a small set of previous events that are responsible for the model's prediction of the target event. In \cite{liudifferential}, the approach involves a smooth parameterization of the GNN
predicted distributions using axiomatic attribution. These distributions are assumed to be on
a low-dimensional manifold. The approach models the distributional evolution as
smooth curves on the manifold and reparameterize families of curves by designing a convex optimization problem. The aim is to find a unique curve that
approximates the distributional evolution and will be useful for human interpretation.

The following ones also design explainers of temporal GNNs but with specific objectives or applications. \cite{vu2022limit} studies the limit of perturbation-based explanation methods. The approach constructs some specific instances of TGNNs and evaluate how reliably node-perturbation, edge-perturbation or both can reliably identify specific graph components carrying out the temporal aggregation in temporal GNNs.
In \cite{yang2023interpretable}, a novel interpretable model on temporal heterogeneous graphs has been proposed. The method constructs
temporal heterogeneous graphs which represent the research interests of the target authors. After the detection task, a deep neural network has been used for the generation process of interpretation on the predicted results. This method has been applied to research interest shift detection of researchers. 
Another related work \cite{han2022interpretable} is on explaining GraphRC which is a special type of GNN and popular because of its training efficiency.  
The proposed method explores the specific role played by each reservoir node (neuron) of GraphRC by using attention mechanism on the distinct temporal patterns in the reservoir nodes.

\subsection{Global Explainers}
\label{sec::global}
Majority of the explainers provide explanation for specific instances and can be seen as \textit{local explainers}. However, global explainers aim to explain the overall behavior of the model by finding common input patterns that explain certain predictions \cite{xgnn}. Global explainers provide a high-level and generic explanation compared to local explainers. However, local explainers can be more accurate compared to global explainers \cite{xgnn} especially for individual instances. We categorize global explainers into following three types.\\
\noindent
\textbf{(1) Generation-based:} These post-hoc methods use either a generator or generative modeling to find explanations. For instance, \textbf{XGNN} \cite{xgnn} uses a reinforcement learning (RL) based graph generator optimized using policy gradient. In contrast, \textbf{GNNInterpreter} \cite{gnninterpreter} is a generative global explainer that maximizes the likelihood of explanation graph being predicted as the target class by the model (the details are in Sec. \ref{sec::generation-based}).\\
\noindent 
\textbf{(2) Concept-based}: These methods provide concept based explanations. Concepts are small higher level units of information that can be interpreted by humans \cite{automatic_concept}. The methods differ in the approaches to find concepts. \textbf{GCExplainer} \cite{global_concept_ex} adapts an image explanation framework known as automated concept based explanation (ACE) \cite{automatic_concept} to find global explanation for graphs. It finds concepts by clustering the embeddings from the last layer of the GNN. A concept and its importance are represented by a cluster and the number of nodes in it respectively. Another method \textbf{GCneuron } \cite{Global_neuron}, which is inspired by Compositional Explanations of Neurons \cite{Neuron-compositional}, finds global explanation for GNNs by finding compositional concepts aligned with neurons. A base concept is a function \(C\) on Graph \(G\) that produces a binary mask over all the input nodes \(V\). A compositional concept is logical combination of base concepts. This method uses beam search to find compositional concept that minimizes the divergence between the concept and the neuron activation. Lastly, \textbf{GLGExplainer} \cite{global_logic} uses local explanations from PGExplainer \cite{pgexplainer} and projects them to a set of learned prototype or concepts (similar to ProtGNN \cite{protgnn}) to derive a concept vector. A concept vector is vector of distances between graph explanation and each prototype. This concept vector is then used to train an Entropy based logic explainable network (E-LEN) \cite{logic-LEN} to match the prediction of the class. The logic formula from the entropy layer for each class acts as explanations.\\
\noindent
\textbf{(3) Counterfactual}: \textbf{Global counterfactual explainer} \cite{Global-counter} finds a candidate set of counterfactuals using vertex re-inforced random walk. It then uses a greedy strategy to select the top $k$ counterfactuals from the candidate set as global explanations. We explain it in more detail in Sec. \ref{sec::Counterfactual-search}. 

\subsection{Causality-based Explainers}
Most of the GNN classifiers learn all statistical correlation between the label and input features. As a result, these may not distinguish between causal and non-causal features and may make prediction using shortcut features \cite{causal-attention}. Shortcut features serve as confounders between the causal features and the prediction. Methods in this category attempt to reduce the confounding effect so that the model exploits causal substructure for prediction and these substructures also act as explanations. They can be categorized into the followings.

\textbf{Self-interpretable methods.} Methods in this category have the explainer architecture inbuilt into the model. 
One of the methods, \textbf{DIR} \cite{D_invariant_rationale} creates multiple interventional distribution by conducting intervention on the training distribution. The invariant part across these distributions is considered as the causal part (see details in Sec. \ref{subsec:: SE- structural-constr}). \textbf{CAL} \cite{causal-attention} uses edge and node attention to estimate causal and shortcut features of the graph. Two classifiers are used to make prediction on causal and shortcut features respectively. Loss on causal features is used as classification loss. Moreover, KL divergence is used to push the prediction based on shortcut features to have uniform distribution across classes. Finally, CAL creates an intervention graph in the representation space via random additions. The loss on this intervened graph classification is considered as the causal loss. These three loss terms are used to reduce the confounding effect and find the causal substructure that acts as explanation. \textbf{DisC} \cite{causal-debiasing} uses a disentangled GNN framework to separate causal and shortcut substructures. It first learns a edge mask generator that divides the input into causal and shortcut substructures. Two separate GNNs are trained to produce disentangled representation of these substructures. Finally, these representations are used to generate unbiased counterfactual samples by randomly permuting the shortcut representation with the causal representation. 

\textbf{Generation-based methods.} These methods use generative modeling to find explanations and are post-hoc. \textbf{GEM~\cite{Gen-causal}} trains a generative auto-encoder by finding the causal contribution of each edge in the computation graph (details are in Sec. \ref{sec::generation-based}). While GEM focuses on the graph space, \textbf{OrphicX} \cite{causal-orphicx} identifies the causal factors in the embedding space. It trains a variational graph autoencoder (VGAE) that has an encoder and a generator. The encoder outputs latent representations of causal and shortcut substructures of input. Generator uses both of these representations to generate the original graph and the causal representation to produce a causal mask on original graph. The information flow between the latent representation of the causal substructure and the prediction is maximized to train the explainer. The causal substructure also acts as an explanation.

\section{Applications}
\label{sec::application}
We describe the explainers methods that are relevant for specific applications in different domains such as in social networks, biology, and computer security.

\paragraph{Computer Security.} This work \cite{he2022illuminati} focuses on designing an explanation framework for cybersecurity applications using GNN models by identifying the important nodes, edges, and attributes that are contributing to the prediction. The applications include code vulnerability detection and smart contract vulnerability detection. Another work \cite{herath2022cfgexplainer} proposes CFGExplainer for GNN oriented malware classification and identifies a subgraph of the malware control flow graph that is most important for the classification. Some other work focuses on the problem of botnet detection. The first method BD-GNNExplainer \cite{zhu2022interpretability} extracts the explainer subgraph by reducing the loss between the classification results generated by the input subgraph and the entire input graph. The XG-BoT detector proposed in \cite{lo2023xg} detects malicious botnet nodes in botnet communication graphs. The explainer is based on the GNNExplainer and saliency map in the XG-BoT. 

\paragraph{Social Networks.}A recent work \cite{rath2021scarlet} studies the problem of detecting fake news spreaders in social networks. The proposed method SCARLET is a user-centric model that uses a GNN with attention mechanism. The attention scores help in computing the importance of the neighbors. The findings include that a person’s decision to spread false information is dependent on its perception (or trust dynamics) of neighbor’s credibility. On the other hand, \textbf{GCAN} \cite{GCAN} uses sequence models. The aim is to find a fake tweet based on the user profile and the sequence of its retweets. The sequence models and GNNs help to learn representation of retweet propagation and representation of user interactions respectively. A co-attention mechanism is further used to learn the correlation between source tweet and retweet propagation and make prediction. In \cite{ma2021understanding}, a GNN model has been proposed along with the explanation of its prediction for the problem on drug abuse in social networks.

\paragraph{Computational Biology.} 
One of the long standing problems in neuroscience is the understanding of Brain networks, especially understanding the Regions of Interests (ROIs) and the connectivity between them. These regions and their connectivity can be modelled as a graph. A recent work on explainability, IBGNN \cite{cui2022interpretable} explores the explainable GNN methods to solve the task of identifying ROIs and their connectivity that are indicative of brain disorders. It uses a perturbation matrix to create an edge mask, and extracts important edges and nodes. A few more works also focus on the same task of identifying ROIs, but use different explanation techniques. In \cite{mauri2022accurate}, the method uses a perturbation matrix with feature masks and optimizes mutual information to find the explanations. This work \cite{zhou2022interpretable} uses Grad-CAM \cite{Excitation-BP} to find important ROIs. \cite{abrate2021counterfactual} uses a search-based method to extract counterfactuals, which can serve as good candidates for important ROIs. The method uses graph edit operations to navigate from input graph to a counterfactual, but it optimizes this by using a lookup database to select edges that are the most effective in discriminating between different predicted classes. As another interesting application, this work \cite{pfeifer2022gnn} explores is related to the extraction of subgraphs in protein-protein interaction (PPI) network, where the downstream task is to detect the relevance of a protein to cancer. 



\paragraph{Chemistry.} GNNs are being used to study molecular properties extensively and often requires explanations to better understand the model's predictions. A recent work \cite{henderson2021improving} focuses on improving self-interpretability of GCNs by imposing orthogonality of node features and sparsity of the GCN's weights using Gini regularization. The intuition behind the orthogonality of features is driven by the assumption that atoms in a molecule can be represented by a linear combination of orthonormal basis of wavefunctions. Another method, APRILE \cite{xu2021aprile} aims at finding the parts of a drug molecule responsible for side effects. It uses perturbation techniques to extract an explanation. In drug design, the method in \cite{jimenez2021coloring} uses integrated gradients \cite{sundararajan2017axiomatic} to assign importance to atoms (nodes) and the atomic properties (node features) to understand the properties of a drug.

\paragraph{Pathology.} In medical diagnosis a challenging task is to understand the reason behind a particular diagnosis, whether it is made by a human or a machine learning system. To this end, explainer frameworks for the machine learning models become useful. In many cases the diagnosis data can be represented by graphs. This work \cite{wu2021counterfactual} builds a graph using words, entities, clauses and sentences extracted from a patient's electronic medical record (EMR). The objective is to extract the entities most relevant for the diagnosis by training an edge mask, and is achieved by minimizing the sum of the elements in the mask matrix. Another method~\cite{jaume2021quantifying} focuses on generating explanations for histology (micro-anatomy) images. It first converts the image into a graph of biological entities, where the nodes could be cells, tissues or some other task specific biological features. Afterwards the standard explainer techniques described in gradient \ref{sec::gradient-based} or perturbation \ref{sec::perturbation} based methods are used to generate the explanations. Another work \cite{yu2021towards} in this field modifies the objective to optimise for both necessity and sufficiency (Sec. \ref{sec::eval}). The explanation is generated in such a way that the mutual information between explanation subgraph and the prediction is maximized. Additionally, the mutual information between the remaining graph after removing the explanation subgraph and the prediction is minimized. 



\section{Datasets}
\label{sec::datasets}

A set of synthetic as well as real-world datasets have been used for evaluating the proposed explainers in several tasks such as node classification and graph classification. Table \ref{tab::dataset} lists down the set of datasets and the corresponding explanation types and tasks used in the literature.
\subsection{Synthetic datasets}
Annotating ground truth explanations in graph data is laborious and requires domain expertise. To overcome this challenge, several explainers have been evaluated using synthetic datasets that are created using certain motifs as ground truth values. We highlight \textit{six} popular synthetic datasets:

\noindent\textbf{BA-Shapes} \cite{ying2019gnnexplainer}: This graph is formed by randomly connecting a base graph to a set of motifs. The base graph is a Barabasi-Albert (BA) graph with $300$ nodes. It includes $80$ house-structured motifs with five nodes each, formed by a top, a middle, and a bottom node type. 
    
\noindent\textbf{BA-Community} \cite{ying2019gnnexplainer}: The BA-community graph is a combination of two BA-Shapes graphs. The features of each node are assigned based on two Gaussian distributions. Also, nodes are assigned a class out of eight classes based on the community they belong to.

\noindent\textbf{Tree Cycle} \cite{ying2019gnnexplainer}: This consists of a 8-level balanced binary tree as a base graph. To this base graph, 80 cycle motifs with six nodes each are randomly connected. It just has two classes; one for the nodes in the base graph and another for nodes in the defined motif.
    
\noindent\textbf{Tree Grids} \cite{ying2019gnnexplainer}: This graph uses same base graph but a different motif set compared to the tree cycle graph. It uses 3 by 3 grid motifs instead of the cycle motifs.
    
\noindent\textbf{BA-2Motifs} \cite{pgexplainer}: This is used for graph classification and has two classes. The base graph is BA graph for both the classes. However, one class has a house-structure motif and another has a 5-node cycle motif.
    
\noindent\textbf{Spurious Motifs} \cite{D_invariant_rationale}: With 18000 graphs in the dataset, each graph is a combination of one base $S$ (Tree, Ladder or Wheel) and one motif $C$ (Cycle, House, Crane). Ground-truth $Y$ is determined by the motif. A spurious relation between $S$ and $Y$ is manually induced. This spurious correlation can be varied based on a parameter that ranges from $0$ to $1$.

\begin{table}[tb]
\vspace{-2mm}
  \centering
  \scriptsize
  \caption{It shows the datasets for different categories, explanation types and tasks. }
    \begin{tabular}{ccccc}
    \toprule
          \textbf{Dataset} & \textbf{References} & \textbf{Nature} & \textbf{Explanation Type} & \textbf{Task}  \\  \midrule
        BA-Shapes & \cite{ying2019gnnexplainer,pgm-ex,RELex,pgexplainer,Gen-causal,cfgnnex,robust-counter}& Synthetic  & Compared to Motif  & Node classification  \\  
        BA-Community & \cite{RL-enhanced,ying2019gnnexplainer, pgexplainer, robust-counter, clear-counter} & Synthetic  & Compared to Motif  & Node classification \\  
       Tree Cycle & \cite{Gen-causal, pgexplainer, RELex, robust-counter,cfgnnex} & Synthetic & Compared to Motif  & Node classification  \\  
        Tree Grids & \cite{ying2019gnnexplainer,pgexplainer, RELex, Gen-causal, robust-counter, cfgnnex} & Synthetic & Compared to Motif  & Node classification \\  
        BA-2Motif & \cite{pgexplainer, subgraphX,GSAT,robust-counter} & Synthetic & Compared to Motif  & Graph classification \\  
        Spurious Motifs & \cite{GSAT,D_invariant_rationale}& Synthetic & Compared to Motif  & Graph classification \\  
        Mutagenicity & \cite{ying2019gnnexplainer, pgexplainer, Gen-causal, protgnn, subgraphX,robust-counter,xgnn} & Real-World & Compared to Chemical property  & Graph classification \\  
        NCI1 & \cite{Gen-causal, robust-counter}& Real-World & Compared to Chemical property  & Graph classification \\  
        BBBP & \cite{protgnn,cf^2-counter, moleculenet} & Real-World & Compared to Chemical property  & Graph classification \\  
        Tox21 & \cite{meg-counter, moleculenet} & Real-World & Compared to Chemical property  & Graph classification \\  
        MNIST-75sp & \cite{pgm-ex,GSAT,mnist_75}& Real-World & Visual  & Graph classification \\  
        Sentiment Graphs & \cite{subgraphX,GSAT,protgnn,sst-datasets} & Real-World & Visual  & Graph classification \\  \bottomrule

    \end{tabular}%
    \vspace{-2mm}  \label{tab::dataset}%
\end{table}%

\subsection{Real-world datasets}
Due to the known chemical properties of the molecules, molecular graph datasets become a good choice for evaluating the generated explanation structure. 
We highlight some widely used molecular datasets for evaluating explainers in the \textit{graph classification task}. 

\noindent\textbf{Mutag} \cite{mutag}: This consists of $4337$ molecules (graphs) with two classes based on the mutagenic effect. Using domain knowledge, specific chemical groups are assigned as ground truth explanations.    
    
\noindent\textbf{NCI1} \cite{NCI1_data}: It is a graph classification dataset with 4110 instances. Each graph is a chemical compound where a node represents an atom and an edge represents a bond between atoms. Each molecule is screened for activity against non-small cell lung cancer or ovarian cancer cell lines.

\noindent\textbf{BBBP} \cite{moleculenet}: Similar to Mutag, Blood-brain barrier penetration (BBBP) is also a molecule classification dataset with two classes with 2039 compounds. Classification is based on their permeability properties.

\noindent\textbf{Tox21} \cite{moleculenet}: This dataset consists of 7831 molecules with 12 different categories of chemical compounds. The categorization is based on the chemical structures and properties of those compounds. 
    
Visual explanation can be an important component of comparing explainers. Hence, researchers also use datasets that do not have ground truth explanations but can be visually evaluated through generated examples. Below are some of the datasets used for visual analysis:

\noindent\textbf{MNIST-75sp}~\cite{mnist_75}: An MNIST image is converted to a super-pixel graph with at most 75 nodes, where each node denotes a ``super pixel''. Pixel intensity and coordinates of their centers of masses are used as the node attributes. Edges are formed based on the spatial distance between the super-pixel centers. Each graph is assigned one of the 10 MNIST classes, i.e., numerical digits. 

\noindent\textbf{Sentiment Graphs}~\cite{sst-datasets}: Graph SST2, Graph SST5, and Graph Twitter are based on text sentiment analysis data of SST2, SST5, and Twitter datasets. A graph is constructed by considering tokens as nodes, relations as edges, and sentence sentiment as its label. The BERT architecture is used to obtain 768-dimensional word embeddings for the dataset. The generated explanation graph can be evaluated for its textual meaning.

\section{Evaluation}
\label{sec::eval}
The evaluation of the explainer methods is based on the quality of the explainer's ability to generate human-intelligible explanations about the model prediction. As this might be subjective depending on the applications in hand, the evaluation measures consider both quantitative and qualitative metrics.

\subsection{Quantitative Evaluation}
Quantitative evaluation metrics help in having a standardized evaluation that is free of human bias. 
For this, explainability is posed as a binary classification problem. The explainers assign a score to the \textit{node features}, \textit{edges}, and \textit{motifs}, which are the most responsible for the prediction according to the explainer. We are also provided with the ground-truth binary labels for the features and structures based, denoting whether they are responsible for the prediction or not. The explainer is then evaluated by comparing these scores to the ground-truth explanation labels using different methods: 
\noindent\textbf{Accuracy \cite{cfgnnex, cf^2-counter}:} To find the accuracy, the top-$k$ edges produced by the explainer are set to be positive, and the rest are negative. These top-k edges and the ground-truth labels are compared to compute the accuracy.

\noindent\textbf{Area Under Curve (AUC) \cite{RELex, agnostic-counter, robust-counter}:} We compare the top-$k$ raw scores directly against the ground-truth labels by computing the area under the ROC curve. 

\noindent\textbf{Fidelity \cite{VGIB, cfgnnex, robust-counter}:} This is used for explainers that generate a subgraph as the explanation. It compares the performances of the base GNN model on the input graph and the explainer subgraph. Let $N$ be the number of samples, $y_i$ is the true label of sample $i$, $\hat{y}_i$ is the predicted label of sample $i$, $\hat{y}^k$ is the predicted label after choosing the subgraph formed by nodes with top-$k$\% nodes, and $\mathbbm{1}[\cdot]$ is the indicator function. Fidelity measures how close the predictions of the explanation sub-graph are to the input graph. For factual explainers, the lower this value, the better is the explanation. It is formally defined as follows:
    \begin{equation*}
        \text{Fidelity} = \frac{1}{N} \sum_{i=1}^{N} {\mathbbm{1}}[y_i = \hat{y}_i] - \mathbbm{1}[y_i = \hat{y}_i^k]
    \end{equation*}    

\noindent\textbf{Sparsity  \cite{VGIB, cfgnnex}:} It measures the conciseness of explanations (e.g., the sub-graphs) that are responsible for the final prediction. Let $|p_i|$ and $|\CG_i|$ denote the number of edges in the explanation, and the same in the original input graph, respectively. The sparsity is then defined as follows:
    \begin{equation*}
        \text{Sparsity} = 1 - \frac{1}{N} \sum_{i=1}^{N} \frac{|p_i|}{|\CG_i|}
    \end{equation*}

\noindent\textbf{Robustness \cite{robust-counter}:} It quantifies how resistant an explainer is to perturbations on input graph. Here perturbations are addition or deletion of edges randomly such that it does not change the prediction of the underlying GNN. The robustness is the percentage of graphs for which these perturbations do not change the explanation.

\noindent \textbf{Probability of Sufficiency (PS) \cite{cf^2-counter, chen2022grease}:} It is the percentage of graphs for which the explanation graph is sufficient to generate the same prediction as the original input graph.

\noindent \textbf{Probability of Necessity (PN) \cite{cf^2-counter, chen2022grease}:} It is the percentage of graphs for which the explanation graph when removed from the original input graph will alter the prediction made by the GNN.


\noindent \textbf{Generalization \cite{RL-enhanced}:} This measures the capability of generalization of the explainer method in an inductive setting. To measure this, the training dataset size is usually varied and the AUC scores are computed for these tests. Generalisation plays an important role in explanability as the good generalizable models are generally sparse in terms of inputs. This metric is highly relevant for the self-interpretable models.



\subsection{Qualitative Evaluation}
Explanations can also be evaluated qualitatively using expert domain knowledge. This mode of evaluation is crucial especially while working with real-world datasets that do not have ground truth labels.

\noindent \textbf{Domain Knowledge \cite{ying2019gnnexplainer}:} Generated explanations can be evaluated for their meaning in the application domain. For example, GNNExplainer~\cite{ying2019gnnexplainer} correctly identifies the carbon ring as well as chemical groups NH2 and NO2, which are known to be mutagenic. 

\noindent \textbf{Manual Scoring \cite{pgm-ex}:} Another method of evaluating the explanations is by asking users (e.g., domain experts) to score, say on a scale of 1-10, the explanations generated by various explainers and compare them. One can also use RMSE scores to quantitatively compare these explainers based on the scores.


\section{Future Directions}
\label{sec:future_work}

\textbf{Combinatorial problems: } Most of the existing explanation frameworks are for prediction tasks such as node and graph classification. However, graphs are prevalent in various domains, such as in social networks~\cite{kempe2003maximizing,medya2020approximate}, healthcare~\cite{wilder2018optimizing}, and infrastructure development~\cite{medya2018noticeable,medya2016towards}. Solving combinatorial optimization problems on graphs is a common requirement in these domains. Several architectures based on Graph Neural Networks (GNNs) \cite{khalil2017learning,manchanda2020gcomb,ranjan2022greed} have been proposed to tackle these problems that are usually computationally hard. However, the explainability of these methods for such combinatorial problems is largely missing. One potential direction is to build frameworks that can explain the behavior of the solution set in such problems. \\
\noindent 
\textbf{Global methods: } 
Most explainers primarily adopt a local perspective by generating examples specific to individual input graphs. From global explanations, we can extract higher-level insights that complement the understanding gained from local explanations (see details on global methods in Sec. \ref{sec::global}). Moreover, global explanations can be easily understood by humans even for large datasets.
Real-world graph datasets often consist of millions of nodes. When generating explanations specific to each instance, the number of explanations increases proportionally with the size of the dataset. As a result, the sheer volume of explanations becomes overwhelming for human cognitive capabilities to process effectively. Global approaches can immensely help in these scenarios. \\
\noindent
\textbf{Visualization and HCI tools: } Graph data, unlike textual and visual data, cannot be perceived by human senses. Thus, qualitative evaluation of explanation becomes a non-trivial problem and often requires expert guidance~\cite{ying2019gnnexplainer,pgm-ex}. This makes crowdsourcing evaluations difficult and not scalable. Other ways to qualitatively assess graph structures for explanation of a certain prediction can be explored. Additionally, since explainability is human-centric, it is crucial that explainers are influenced by human cognition and behavior, particularly those of domain experts~\cite{liao2021human} while using GNNs in making important decisions \cite{medya2022exploratory}.
 HCI research can help in designing the interface for the experts to assess the generated explanation graphs~\cite{cfgnnex}. \\
\textbf{Temporal GNNs: } Temporal graph models are designed to predict the graph structure and labels in the future by exploiting how the graph has evolved in the past. This increases the complexity of explanations significantly as they now involve combinations of graph structures at different time intervals. Existing methods~\cite{fan2021gcn,xie2022explaining,he2022explainer,kosan2021event} mostly focus on discrete-time models where graphs are provided at different points in time. Future works can explore ways to explain the prediction of a continuous-time dynamic graph model, where interactions happen in real time~\cite{xiaexplaining}. One direction could be to optimize over a parameterized temporal point process~\cite{trivedi2019dyrep}. 
\section{Conclusions}
\label{sec:conclusion}
In this survey, we have provided a comprehensive overview of explanation methods for Graph Neural Networks (GNNs). Besides outlining some background on GNNs and explainability, we have presented a detailed taxonomy of the papers from the literature. By categorizing and discussing these methods, we have highlighted their strengths, limitations, and applications in understanding GNN predictions. Moreover, we have highlighted some widely used datasets and evaluation metrics in assessing the explainability of GNNs. As GNNs continue to play a significant role in various fields, such as healthcare, recommendation systems, and natural language processing, the need for interpretable and transparent models becomes increasingly important. Overall, we believe this survey serves as a valuable resource for researchers and practitioners interested in the explainability of GNNs and provides a foundation for further advancements in interpretable graph representation learning. 


\bibliographystyle{plain}
\bibliography{arxiv_main}

\end{document}